# Exact Optimality Certification of QPLIB 2096 through Complete Set Partitioning

Jingyang Du[1,†], Zixiong Yi[†], Yang Shen[2,3,*]

[1]School of Journalism and Communication, Minzu University of China, Beijing, China
[2]School of Journalism and Communication, Tsinghua University, Beijing, China
[3]College of AI, Tsinghua University, Beijing, China

[†]These authors contributed equally to this work and share first authorship.
[*]Corresponding author: Yang Shen (124739259@qq.com)
ORCID: Yang Shen 0000-0003-4814-9018

**Abstract.** We present an exact computational proof that the frozen QPLIB_2096 instance has global minimum 7,025,720. The solution and certification approach was explored and refined through ZiYor, a self-evolving AI system. The authors provide the mathematical account and exact validation reported here. The original formulation has 300 binary variables, 6,900 triangle inequalities, and 25 nonconvex quadratic constraints. A coefficient-level reconstruction identifies its variables with the unordered pairs of 25 items. The triangle inequalities encode partitions, while each quadratic constraint becomes a bound on the total weight of edges incident to one partition block. This yields a set-partitioning formulation over all capacity-feasible blocks, with a separate edge-cost graph determining the objective. Exact enumeration examines 245,505 subsets of sizes one through six and finds 14,981 feasible blocks. Every six-element subset is infeasible; nonnegative capacity weights exclude all larger blocks. An integer price vector satisfies the dual inequality for every feasible block and has sum 2,229,456. Subtracting this sum from the total objective edge cost 9,255,176 gives a lower bound of 7,025,720, attained by the supplied six-block partition. Fresh solver-free checks validate the candidate against both frozen model formats and verify the certificate. The result is an instance-specific optimality proof with an exact primal and dual witness; no cross-instance performance conclusion is drawn.

## 1 Introduction

A feasible solution and a proof of optimality answer different questions in nonconvex discrete optimization. A feasible point establishes an upper bound for a minimization problem. A global claim additionally requires a lower bound valid for every feasible point. When a model has exploitable combinatorial structure, that lower bound may admit a much smaller witness than the numerical search that first produced the solution. This paper develops that distinction for the frozen QPLIB_2096 instance and documents an exact certificate that can be checked without an optimization solver.

QPLIB provides a common collection of quadratic optimization instances, model formats, and solution points [1]. The QPLIB_2096 instance is classified as LBQ: its objective is linear, its variables are binary, and its constraints include nonconvex quadratic expressions [2]. Its apparent dimension is 300 binary variables, but these variables encode all unordered pairs of 25 items. The resulting partition structure makes a complete feasible-block representation practical for this instance.

The numerical result studied here is already reflected in the public benchmark. On 13 October 2026, the instance page displayed objective 7,025,720 and zero reported infeasibility [2]. The QPLIB homepage records an update dated 5 October 2026 and credits ZiYor with the improved solution and an exact optimality certificate [3]. We preserve this public attribution. The paper provides a mathematical account and fresh validation of the supplied result; the public website is evidence of benchmark status, while the original-model checks and integer inequalities provide the optimality argument.

The main result combines three verifiable ingredients. First, coefficient-level checks bind the pairwise partition interpretation to every constraint of the frozen model. Second, complete enumeration identifies the entire admissible block family, including a finite argument excluding larger blocks. Third, a 25-entry integer price vector bounds the internal objective gain of every admissible block. A feasible partition attains this bound, certifying both the original optimum and the tightness of the associated set-partitioning linear relaxation.

### 1.1 Relation to existing methods

The reformulation follows the established set-partitioning principle: select admissible subsets so that each item is covered exactly once. Its price inequalities are a direct application of linear-programming weak duality. The separation of numerical search from exact bound validation also has clear precedent. Held, Cook, and Sewell construct safe bounds from numerical information in their work on maximum-weight stable sets and graph coloring [4]. Exact rational linear-programming approaches provide another route to reliable conclusions from numerical optimization [5]. These precedents are essential to the scope of the contribution: the present result concerns the explicit structural reduction and complete certificate for one frozen benchmark instance.

The supplied discovery implementation uses SciPy's mixed-integer interface with HiGHS [6,7]. Its optimization phase can suggest a partition, but a floating-point solver status does not establish the certificate in this paper. Acceptance instead checks original coefficients, exact candidate values, complete block coverage, and integer inequalities. The fresh work reported here is verification and mathematical reconstruction; the original optimization search was not rerun.

## 2 Original model and recovered structure

Let $V = \{1, ..., 25\}$. For each unordered pair $\{i, j\}$, let $x_{ij}$ be binary, with symmetric notation $x_{ji} = x_{ij}$ and $x_{ii} = 0$. Two symmetric nonnegative integer arrays have different roles: $c_{ij}$ determines objective cost, and $w_{ij}$ determines capacity. Missing edge coefficients are zero. The model recovered from the frozen files is

$$\text{minimize } f(x) = \sum_{i<j} c_{ij} x_{ij} \tag{1}$$

$$x_{ij} \le x_{ik} + x_{kj} \quad \text{for distinct } i, j, k, \qquad x_{ij} \in \{0, 1\} \tag{2}$$

$$\sum_{j<k;\, j,k\neq i} w_{jk} x_{ij} x_{ik} \ge q \quad \text{for each } i \in V, \quad q = 83{,}748 \tag{3}$$

There are three distinct inequalities in (2) for each unordered triple, giving 3 × C(25,3) = 6,900 linear constraints. Equation (3) contributes one quadratic row per item. These identities were checked against the full coefficient lists, rather than inferred from dimension counts. The variable map is recovered from the supports of the quadratic rows: every variable belongs to exactly two 24-variable stars, which determine its endpoints. The check then verifies the bijection onto all 300 unordered pairs, the complete triangle family, and every term of all 25 quadratic polynomials.

Table 1. Exact properties of the frozen instance.

| Quantity | Value | Meaning |
|---|---|---|
| Items and pair variables | 25; 300 | One binary variable for each unordered pair |
| Constraints | 6,925 | 6,900 linear triangles and 25 quadratic rows |
| Nonzero objective coefficients | 295 | Five pair costs are zero |
| Nonzero capacity weights | 295 | All capacity weights are nonnegative |
| Total capacity weight W | 124,624 | Each unordered edge counted once |
| Quadratic lower bound q | 83,748 | Common right-hand side of (3) |
| Incident capacity H = W - q | 40,876 | Upper bound for each partition block |
| Total objective edge cost C | 9,255,176 | Zero original objective constant |

The two edge arrays must remain separate. A block can be capacity-feasible because of its $w$-weights and still have an objective contribution governed by different $c$-weights. Reusing capacity gains as objective gains would solve a different problem. In the supplied implementation, feasible blocks are first enumerated using the capacity graph and then reweighted using the objective graph.

# 3 Equivalence with complete set partitioning

## 3.1 From binary distances to partition blocks

**Proposition 1.** Binary vectors satisfying all triangle inequalities in (2) are in one-to-one correspondence with partitions of $V$, where $x_{ij} = 0$ if and only if items $i$ and $j$ are in the same block.

**Proof.** Define $i \sim j$ when $i = j$ or $x_{ij} = 0$. Reflexivity and symmetry follow from the definition. If $x_{ij} = x_{jk} = 0$, then the triangle inequality gives $x_{ik} \le 0$, so the relation is transitive. Its equivalence classes form a partition. Conversely, assign zero within each block and one between different blocks. If $x_{ij} = 1$, at least one of $x_{ik}$ and $x_{kj}$ is one; if $x_{ij} = 0$, the inequality is immediate. Thus every partition satisfies all triangles, and the two constructions are inverse.

For a nonempty block $B$, define its incident capacity weight $h(B)$ and its internal objective gain $g(B)$ by

$$h(B) = \sum_{i<j;\ \{i,j\}\cap B\neq\varnothing} w_{ij}, \quad g(B) = \sum_{i<j;\ i,j\in B} c_{ij} \tag{4}$$

The incident quantity includes both edges internal to $B$ and edges crossing its boundary, each counted once. It is neither a vertex-weight sum nor just a cut weight. Let $W$ denote the total capacity edge weight. For any item $i$ in block $B$, the product $x_{ij}x_{ik}$ equals one exactly when both $j$ and $k$ lie outside $B$. Therefore the left side of (3) is the weight of all edges wholly outside $B$, equal to $W - h(B)$.

**Proposition 2.** A partition satisfies the original quadratic constraints if and only if every block satisfies $h(B) \le H$, where $H = W - q = 40{,}876$. Its objective equals total edge cost minus total internal block gain.

$$\begin{gathered} F = \{B \subseteq V : B \neq \varnothing,\ h(B) \le H\}, \\ f(P) = C - \sum_{B\in P} g(B) \end{gathered} \tag{5}$$

**Proof.** Substituting $W - h(B)$ into (3) gives exactly the stated capacity inequality. This holds for every item in the same block. For the objective, every unordered edge is either internal to one block or joins two different blocks. Only the latter contributes to (1). Subtracting all internal costs from $C$ therefore gives the original objective. Together with Proposition 1, this proves equivalence in both directions.

Figure 1 shows the supplied six-block witness and the equivalent model representations. The witness has one block of size five and five blocks of size four; Table 3 gives their complete memberships and exact quantities.

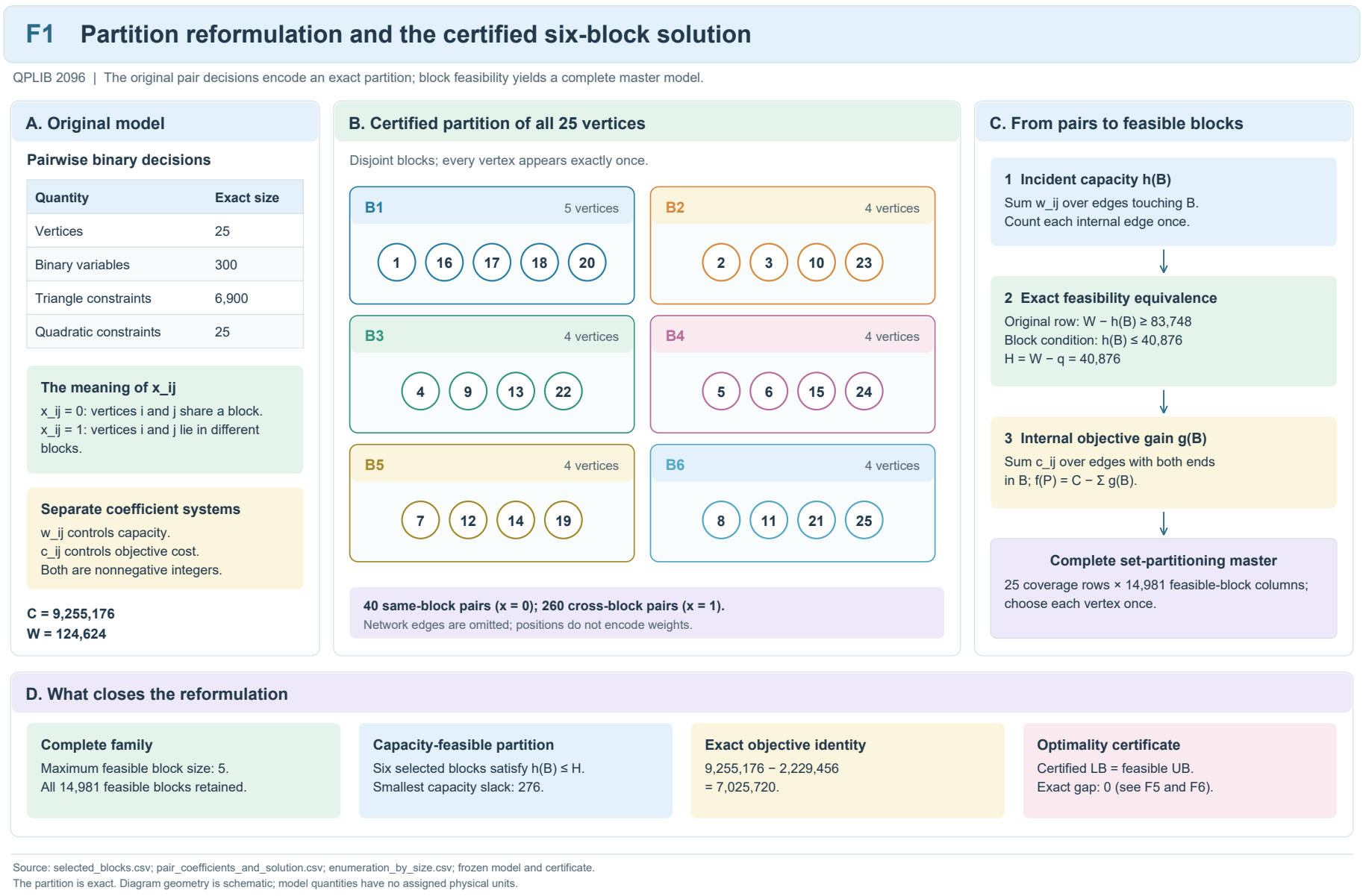

Figure 1. Partition reformulation and the certified solution. (A) Original model dimensions and distinct coefficient systems. (B) All 25 vertices in six disjoint blocks. (C) Block feasibility and the complete set-partitioning model. (D) Completeness, feasibility, objective identity, and certificate closure. Node positions are schematic; Table 3 records exact block quantities.

### 3.2 Complete master problem and enumeration

Introduce one binary variable $y_B$ for every block in $F$. The equivalent complete master problem maximizes internal gain:

$$\begin{aligned} &\text{maximize } \textstyle\sum_{B\in F} g(B)y_B \\ &\text{subject to } \textstyle\sum_{B\in F;\, i\in B} y_B = 1 \quad \text{for all } i \in V,\ \ y_B \in \{0,1\} \end{aligned} \tag{6}$$

The producer enumerates feasible blocks recursively in increasing item order. If $d_v = \sum_{u\neq v} w_{uv}$, adding a new item $v$ to a block updates its incident weight as

$$h(B \cup \{v\}) = h(B) + d_v - \textstyle\sum_{u\in B} w_{uv},\quad v \notin B \tag{7}$$

The subtracted terms are edges already included in $h(B)$. Nonnegative weights imply that $h$ is monotone under inclusion, so an infeasible prefix can be pruned along with every extension. Increasing item order visits each retained subset once. A separate checker deliberately avoids this recurrence: it computes each subset's incident weight by scanning the original edge list.

For this instance, the checker enumerates every subset of sizes one through six. Table 2 reports exact counts. The minimum incident weight among all 177,100 six-element subsets is 47,232, exceeding the capacity by 6,356. Every larger set contains a six-element subset, and monotonicity then excludes it. Thus the 14,981 retained blocks form the complete family $F$, with maximum feasible block size exactly five.

Table 2. Exhaustive verification of the feasible block family.

| Block size | Subsets checked | Feasible blocks | Minimum price slack |
|---|---|---|---|
| 1 | 25 | 25 | 29,278 |
| 2 | 300 | 300 | 52,852 |
| 3 | 2,300 | 2,300 | 26,402 |
| 4 | 12,650 | 12,249 | 0 |
| 5 | 53,130 | 107 | 0 |
| 6 | 177,100 | 0 | Not applicable |
| Total | 245,505 | 14,981 | 0 over feasible blocks |

Figure 2 summarizes all examined and capacity-feasible subset counts, the minimum incident weight at each size, and the feasible fraction. The exact counts are also reported in Table 2.

Complete enumeration has exponential worst-case size. The 25-item interpretation does not make the general problem polynomial, and the original $2^{300}$ binary encodings are not examined individually. The concrete benefit is that all admissible columns can be materialized here. Using a direct edge scan, capacity verification takes $O(M\sum_{k=1}^{6}C(n,k))$ integer additions and comparisons, where $M$ is the number of nonzero capacity edges; gain evaluation adds $O(\sum_{B\in F}|B|^2)$ operations. These are arithmetic-operation bounds. Bit costs depend on coefficient size, and neither this bound nor the observed block count gives a general bound on mixed-integer search time.

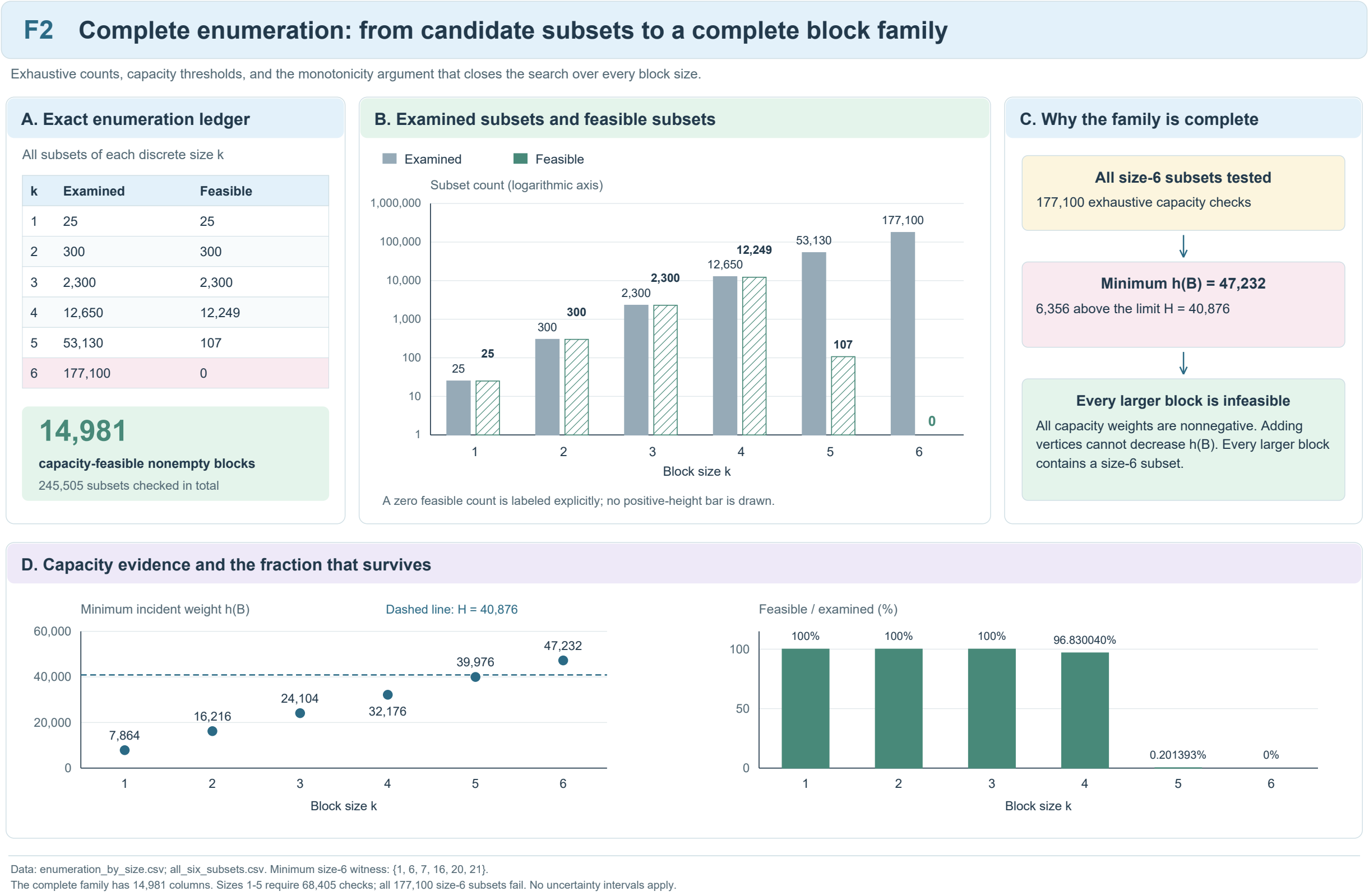

Figure 2. Complete enumeration and the maximum feasible block size. (A) Exact counts. (B) Examined and feasible subsets on a logarithmic axis, with zero explicitly labeled. (C) Size-six exclusion and nonnegative-weight monotonicity. (D) Minimum incident weight and feasible fractions. The complete family contains 14,981 blocks; the numerical ledger is also given in Table 2.

# 4 An exact integer certificate

## 4.1 Price inequalities and the lower bound

Relax the binary variables in (6) to nonnegative real variables. Upper bounds of one are redundant because every nonempty block appears in at least one equality with nonnegative coefficients. Its linear-programming dual minimizes the sum of item prices, subject to one inequality per feasible block:

$$\begin{aligned} &\text{minimize } \textstyle\sum_{i\in V} p_i \\ &\text{subject to } \textstyle\sum_{i\in B} p_i \geq g(B) \quad \text{for every } B \in F \end{aligned} \tag{8}$$

The equality constraints make prices unrestricted in the formal dual. In this instance every singleton is feasible and has zero gain, so the singleton inequalities additionally imply nonnegative prices. For acceptance, integrality of the supplied prices is convenient but not necessary for the general dual argument.

**Theorem 1.** Suppose a price vector satisfies every inequality in (8) over the complete feasible block family. Then every feasible original-model vector has objective at least $C - \sum_{i\in V} p_i$. If an original-model feasible vector attains this value, it is globally optimal.

**Proof.** By Propositions 1 and 2, every feasible original vector corresponds to a partition $P$ of feasible blocks. Summing the price inequality over those blocks gives

$$\sum_{B\in P} g(B) \leq \sum_{B\in P} \sum_{i\in B} p_i = \sum_{i\in V} p_i \tag{9}$$

The last equality holds because the blocks cover every item exactly once. Substitution into (5) yields the lower bound. Equality with the value of an independently feasible original vector closes the minimization bound and proves optimality.

The supplied integer vector is listed in Appendix A. Its sum is 2,229,456. Exact checking finds no negative price slack among the 14,981 feasible blocks, a minimum slack of zero, and 25 tight blocks. Consequently,

$$f(x) \geq 9{,}255{,}176 - 2{,}229{,}456 = 7{,}025{,}720 \quad (10)$$

### 4.2 Candidate partition and equality

The supplied candidate decodes to the six blocks in Table 3. Each is capacity-feasible and price-tight. Their internal gains sum to 2,229,456, exactly the certified upper bound on total internal gain. The reconstructed pair vector contains 40 zero variables and 260 one variables; all 300 values were provided explicitly. Direct evaluation of every original constraint and the original objective accepts the candidate at 7,025,720.

Table 3. Candidate partition and exact block quantities. Each block's price sum equals its internal gain.

| Members | Incident weight h(B) | Capacity slack H - h(B) | Gain g(B) |
|---|---|---|---|
| {1, 16, 17, 18, 20} | 40,600 | 276 | 418,112 |
| {2, 3, 10, 23} | 37,640 | 3,236 | 358,488 |
| {4, 9, 13, 22} | 38,224 | 2,652 | 415,616 |
| {5, 6, 15, 24} | 36,016 | 4,860 | 278,288 |
| {7, 12, 14, 19} | 38,728 | 2,148 | 437,176 |
| {8, 11, 21, 25} | 35,208 | 5,668 | 321,776 |

The incident weights and capacity slacks of the six selected blocks are shown in Figure 3. Their positive capacity slacks establish feasibility of the corresponding quadratic rows.

Theorem 1 therefore establishes the global minimum of the frozen original instance. It also yields a stronger statement about this particular reformulation: the feasible integral cover and the feasible linear-programming dual have the same gain value, 2,229,456. By weak duality, the complete master problem's linear relaxation attains that value too, so its integrality gap is zero for this instance. No general integrality property of the set-partitioning formulation follows.

Figure 4 compares each selected block's internal gain with its price sum and displays all 25 integer prices. The paired block quantities coincide exactly, and both total gain and total price equal 2,229,456. Table A1 retains the complete numerical witness.

### 4.3 Numerical discovery and exact repair

A numerical price vector may guide certificate construction. One sufficient repair procedure is to round each finite price upward to an integer vector $r$, compute the largest positive integer deficit over the complete block family, and add that deficit to every component:

$$\delta = \max(0, \max_{B \in F}\{g(B) - \textstyle\sum_{i \in B} r_i\}), \quad p_i = r_i + \delta \quad (11)$$

Every block is nonempty, so this addition increases its price sum by at least $\delta$ and repairs every deficit. The supplied certificate records zero added repair per price. Our acceptance claim relies on the final integer inequalities, not on reproducing the earlier numerical price computation or trusting its rounding metadata. A repaired feasible vector can yield a weaker bound; exact closure must still be demonstrated against a feasible candidate.

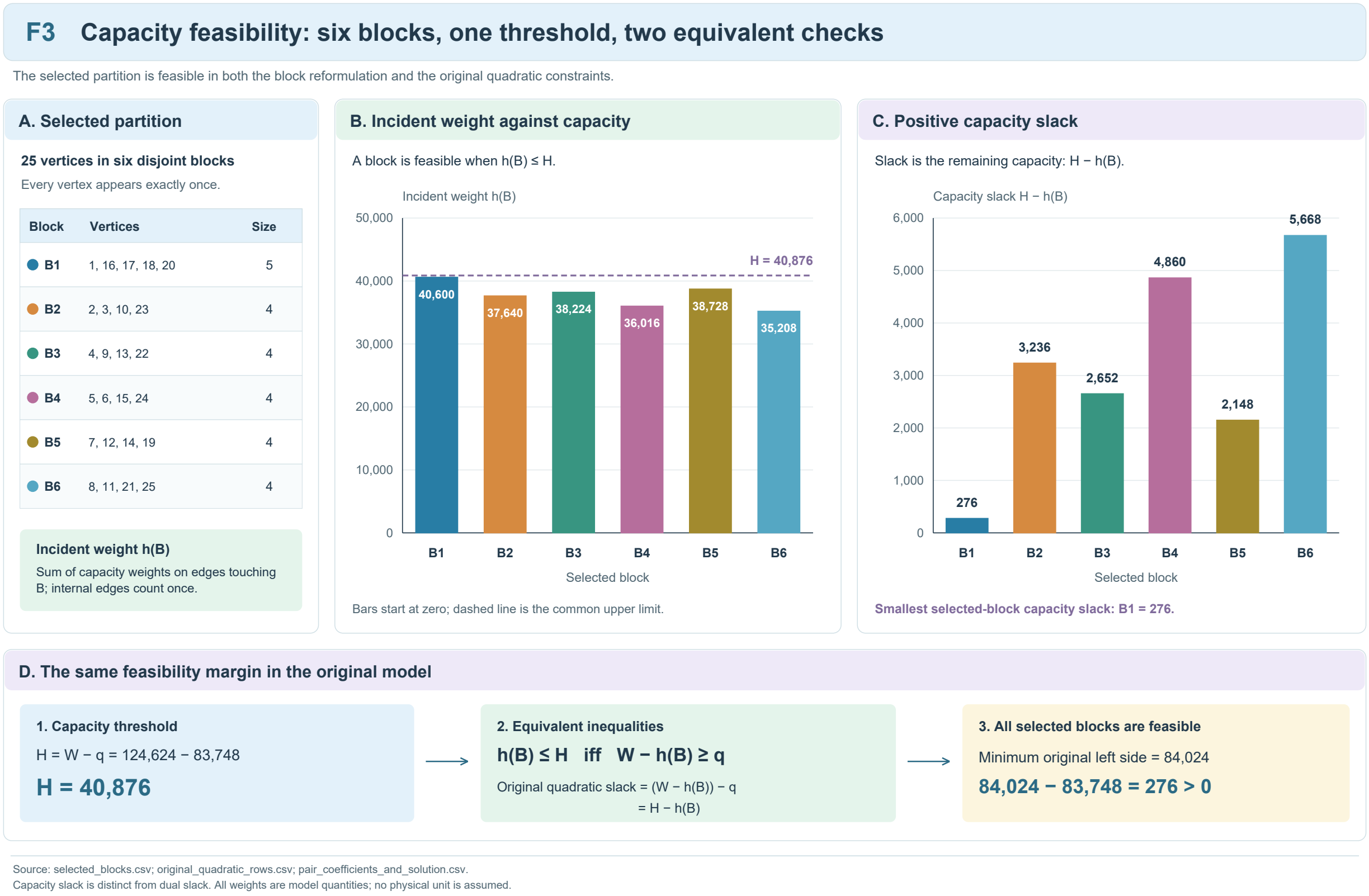

Figure 3. Capacity feasibility of the selected partition. (A) Block memberships. (B) Incident weights against H=40,876. (C) Positive capacity slacks, with selected-block minimum 276. (D) Equivalence with the original quadratic inequalities. Capacity slack and dual slack are different quantities; every selected block has zero dual slack.

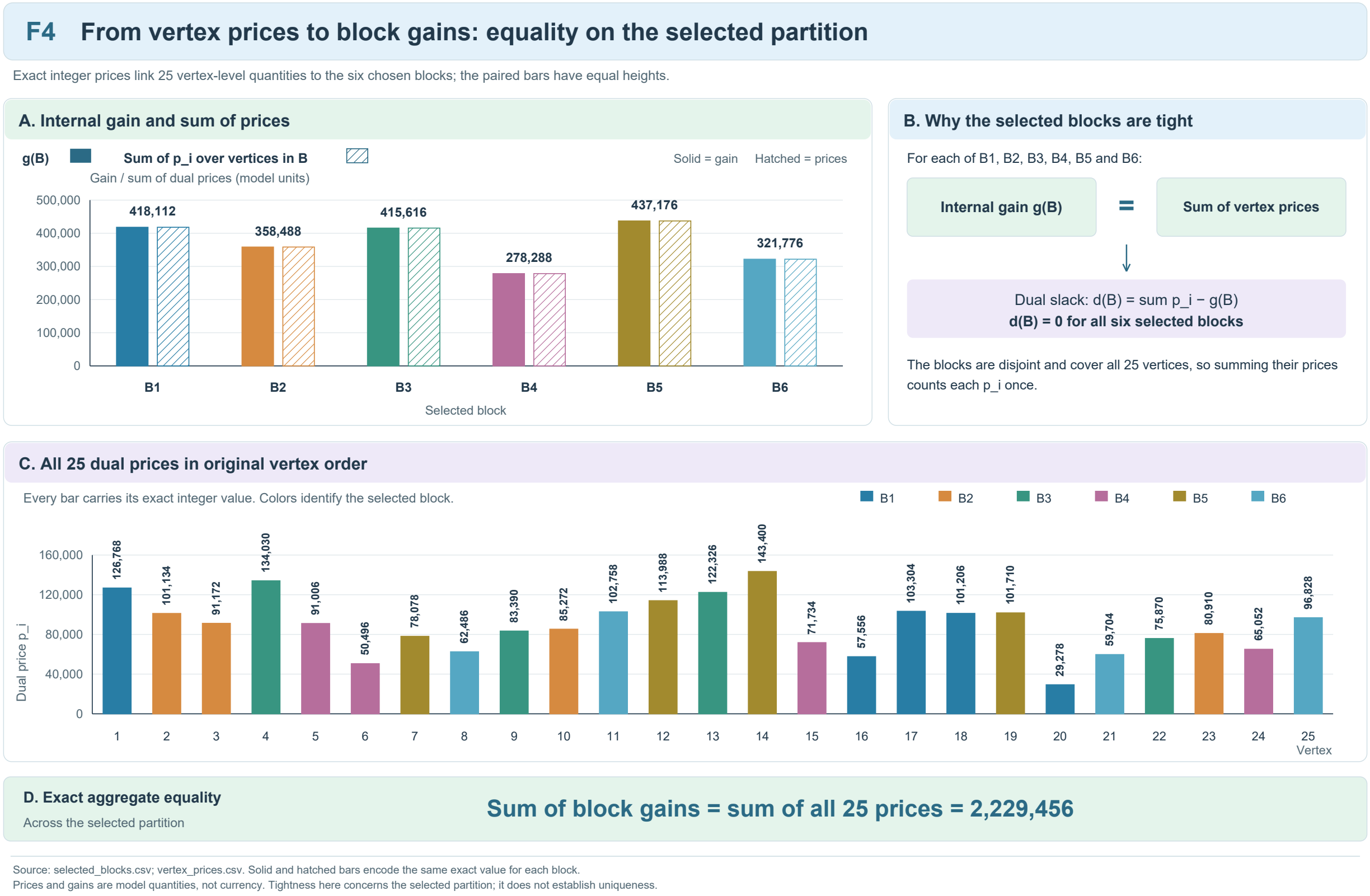

Figure 4. Equality between gains and prices. (A) Equal-height paired bars for all six selected blocks. (B) Their zero dual slacks. (C) All 25 integer prices in vertex order. (D) Total gain equals total price, 2,229,456. Colors identify block membership. Table A1 gives the full price witness; values are model units.

## 5 Computational validation and benchmark context

### 5.1 Frozen inputs and fresh validation

The principal input was the supplied results archive QPLIB_2096_7025720_results_package.zip. Its SHA-256 matched the accompanying checksum, and its internal file checksums were verified. It contains the original QPLIB and AMPL MOD representations, a complete solution vector, the integer certificate, acceptance checkers, and the discovery sources. Full identities for the model, candidate, and certificate are recorded in Appendix B and the accompanying audit files. Checksums establish byte identity; the mathematical checks establish feasibility and bound validity.

Four solver-free acceptance commands were freshly executed for this manuscript on 13 October 2026. Table 4 separates their purposes. The format-identity checker compares both frozen encodings at coefficient level. The candidate checkers evaluate objective, constraints, variable bounds, and integrality. The certificate checker reconstructs the partition model from the MOD file, enumerates the complete block family, and checks exact prices. All four processes returned success. The fresh verification environment was Python 3.12.14 on Windows 11. Exact rational evaluation in the QPLIB checker includes a rational $10^{-9}$ acceptance tolerance, but the observed maximum constraint, bound, and integrality violations were all exactly zero. The MOD checker uses zero-tolerance acceptance, and the additional raw-MOD reconstruction independently obtained the same exact objective and zero violations.

Table 4. Acceptance stages executed for this manuscript.

| Stage | Objects checked | Result |
|---|---|---|
| Format identity | 300 variables, objective, all 6,925 constraints and domains across QPLIB and MOD | Passed |
| Original QPLIB candidate | Complete candidate against original coefficients, bounds and integer domains | Passed |
| Original MOD candidate | Exact original objective and every original constraint | Passed |
| Integer certificate | Model reconstruction, 245,505 subsets and all 14,981 price inequalities | Passed |

These checks have a narrower scope than reproducing the original optimization run. The discovery sources record a complete-pool mixed-integer approach and SciPy/HiGHS dependencies; they were inspected but no optimizer was called in this study. Historical environment metadata names an Intel Core i7-11700F system, 32 GiB memory, SciPy 1.17.1 and HiGHS 1.12.0. Those are archived discovery metadata, not a newly reproduced performance configuration. We report no discovery-time speedup, success probability, or comparative runtime.

Independence also has a specific meaning. The certificate checker does not load the producer's block pool, its numerical optimizer, or a previous verification result. It independently enumerates subsets by direct incident-edge scans. Some supplied checkers share a MOD parser, so separate executable checks alone are not fully independent implementations. The original QPLIB representation, full cross-format identity test, and an additional narrow raw-MOD reconstruction provide complementary checks, but do not constitute a formally verified proof assistant.

### 5.2 Objective comparisons and public status on 13 October 2026

The archived reports record an earlier public reference of 7,064,664 and a local baseline of 7,404,632. Relative to the earlier public reference, the certified objective is lower by 38,944, or 0.551251%. Relative to the recorded local baseline, it is lower by 378,912, or 5.117229%. These comparisons use different denominators and describe objective values; they do not measure runtime or general solver quality.

Table 5. Objective-value comparisons for the minimization instance.

| Reference | Reference objective | Decrease to 7,025,720 | Relative decrease |
|---|---|---|---|
| Archived local baseline | 7,404,632 | 378,912 | 5.117229% |
| Archived earlier public reference | 7,064,664 | 38,944 | 0.551251% |
| Public instance page observation (13 October 2026) | 7,025,720 | 0 | 0% |

Figure 5 combines the historical objective references with primal-dual bound closure and the complete dual-slack distribution. Its four objective categories are archived observations and the verified candidate, not synchronized trials of different algorithms. The 25 tight blocks include all six blocks of the candidate partition.

The supplied official-update note dated 6 October described a discrepancy between the homepage update and the numerical instance page. The instance page observed on 13 October 2026 reported 7,025,720 [2], consistent with the supplied candidate's objective. The homepage observed on that date attributed the update to ZiYor [3]. This distinction matters for chronology: the 38,944 improvement is relative to the archived earlier reference, not an improvement over the published value observed on 13 October 2026. Attribution of the public contribution is retained separately from the authorship of this manuscript. A direct comparison of the public solution saved on 13 October 2026 and the local candidate also found identical values for all 300 original variables after interpreting omitted public entries as zero. The files have different hashes because their formatting and explicit-zero conventions differ.

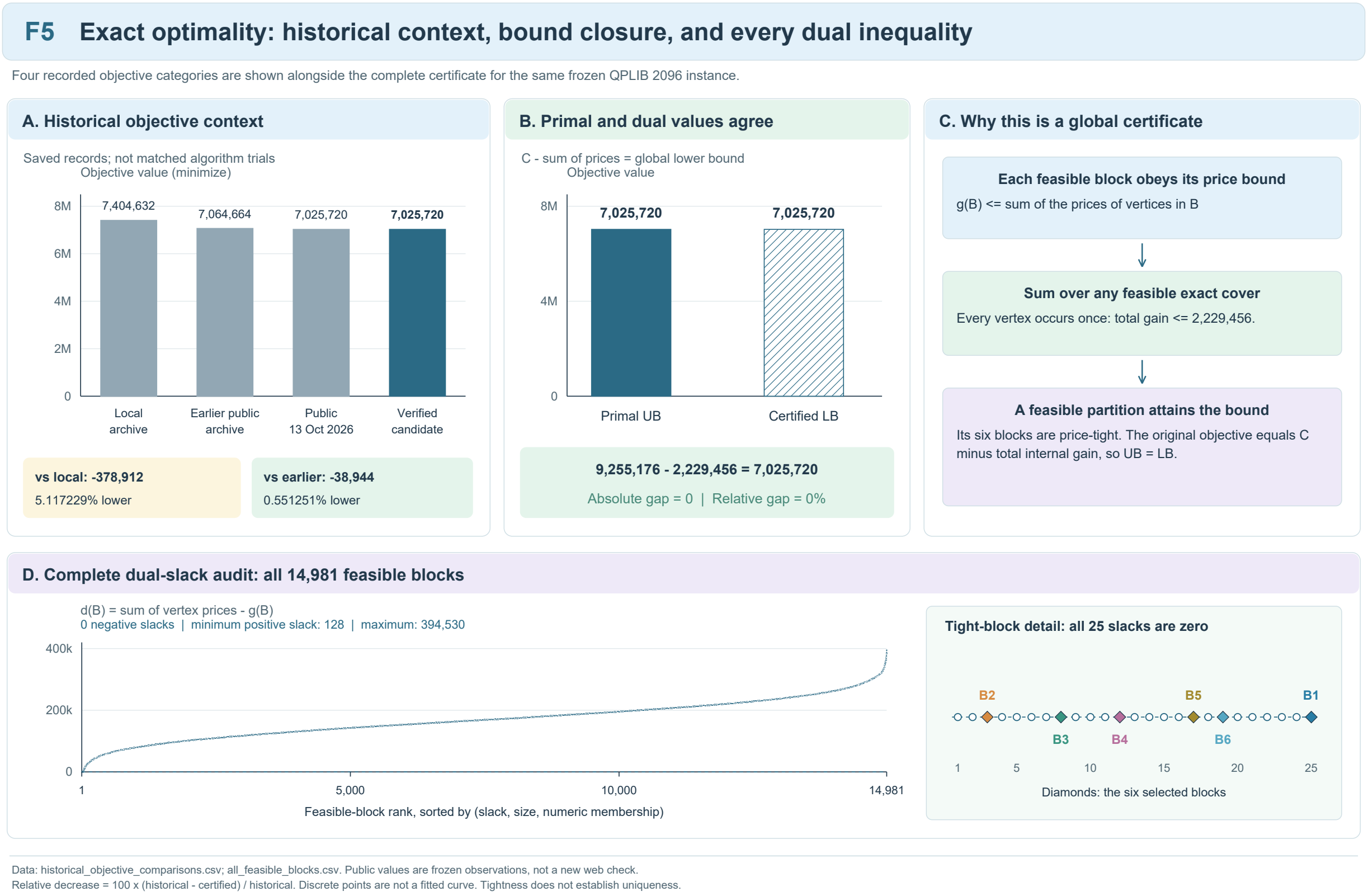

Figure 5. Exact optimality and historical context. (A) Archived objectives, the saved public observation of 13 October 2026, and the verified candidate. (B) UB=LB=7,025,720. (C) The global-bound argument. (D) All 14,981 dual slacks and the 25 tight blocks, including six selected blocks. Historical categories are not synchronized solver trials; tightness does not establish uniqueness.

## 6 Discussion and limitations

The proof is compact in its final numerical witness: 25 integer prices and one feasible partition. Its credibility nevertheless depends on a larger validation chain. The admissible block family must be complete, the block model must be equivalent to the frozen original model, and the candidate must be feasible in that original model. Checking price inequalities over an incomplete pool would bound only a restricted master problem and could not certify the original optimum.

The exclusion of six-element blocks is especially useful here because it turns a finite subset audit into coverage of all block sizes. Nonnegative capacity weights are essential to this step. Negative weights, changes to the quadratic supports, a different triangle system, or modified coefficients can invalidate the argument. The certificate must therefore be rebound and rechecked for any changed model. The maximum size of five and the zero integrality gap are properties of this instance.

The numerical evidence concerns one benchmark instance. The archived reports mention other QPLIB cases, but their raw validation materials were not part of the same complete certificate chain and they are not treated as additional experimental samples. Broad algorithm comparisons would require a defined instance suite, matched budgets, reproducible configurations, and appropriate treatment of unsuccessful runs. No such comparisons are claimed here.

The approach suggests a useful direction for further work: characterize when an admissible-block relaxation is strong enough to admit a small exact dual witness, and how to verify coverage when the full block family is too large to enumerate. Pricing or separation procedures could then replace explicit enumeration, but their exactness and completeness would need their own proof. The present result establishes a fully enumerable case that can serve as a reference for that investigation.

## 7 Conclusion

For the frozen QPLIB_2096 instance, the original binary quadratically constrained formulation is exactly equivalent to a complete set-partitioning problem on 25 items. Exhaustive enumeration and monotonicity establish a family of 14,981 feasible blocks. A 25-entry integer dual vector bounds total internal gain by 2,229,456, and a feasible six-block partition attains this gain. The resulting original-model lower and upper bounds coincide at 7,025,720. Fresh solver-free validation confirms the proof chain, while the QPLIB page observed on 13 October 2026 independently recorded the same solution value. The contribution is a precise, reproducible account of this instance's exact optimum.

## Appendix A Complete integer price witness

Table A1 gives every price in the item numbering recovered from the original quadratic-row supports. The sum is 2,229,456. The vector and all block inequalities are checked with exact arithmetic. Item labels refer to this recovered map, not to a lexicographic assumption about the original variable names.

Table A1. Integer item prices.

| Item | Price | Item | Price | Item | Price | Item | Price | Item | Price |
|---|---|---|---|---|---|---|---|---|---|
| 1 | 126,768 | 6 | 50,496 | 11 | 102,758 | 16 | 57,556 | 21 | 59,704 |
| 2 | 101,134 | 7 | 78,078 | 12 | 113,988 | 17 | 103,304 | 22 | 75,870 |
| 3 | 91,172 | 8 | 62,486 | 13 | 122,326 | 18 | 101,206 | 23 | 80,910 |
| 4 | 134,030 | 9 | 83,390 | 14 | 143,400 | 19 | 101,710 | 24 | 65,052 |
| 5 | 91,006 | 10 | 85,272 | 15 | 71,734 | 20 | 29,278 | 25 | 96,828 |

# Appendix B Reproducibility and file identity

The verification workflow operates on the frozen model and certificate files. It first checks archive and internal checksums, then validates format identity, original-model candidate feasibility, and the complete integer certificate. The original results archive retains the exact checkers and discovery sources. The manuscript delivery includes the fresh audit results and an additional reconstruction script. No public repository or persistent identifier is assigned to this local delivery.

Figure 6 shows the three verification branches and how they establish original-model feasibility, completeness of the feasible block family, and validity of the global lower bound. Table 4 lists the acceptance stages.

**Original results archive SHA-256**
094e1bb152736633f60322e74c2b112c93b3e4720e89d23cabd92c1c566fe9f3

**Candidate solution SHA-256**
d26b52610b9a00ce6b87678e0382a347934b19e02097bfe243f17416f7a153e1

**Frozen MOD SHA-256**
60316999cbf50e07fc40d73e730b06f8d436e4c4e4379097b8e1b117ef399696

From the accompanying bundle root, run **python verify_results.py --out audit-rerun**. This entry point verifies the packaged source checksums and executes the four original acceptance checks in fresh processes without calling an optimizer. It refuses to reuse an existing output directory. The bundled independent reconstruction can also be run separately; its output has a distinct identity from the original acceptance checks.

Original-model verification and complete block enumeration use only the Python standard library. The optional discovery path requires the SciPy/HiGHS stack recorded in the archived environment. A successful verification run does not imply that the discovery run has been reproduced. The locally supplied historical report is preserved as provenance; its earlier website status and its block-size error are not used as current mathematical facts.

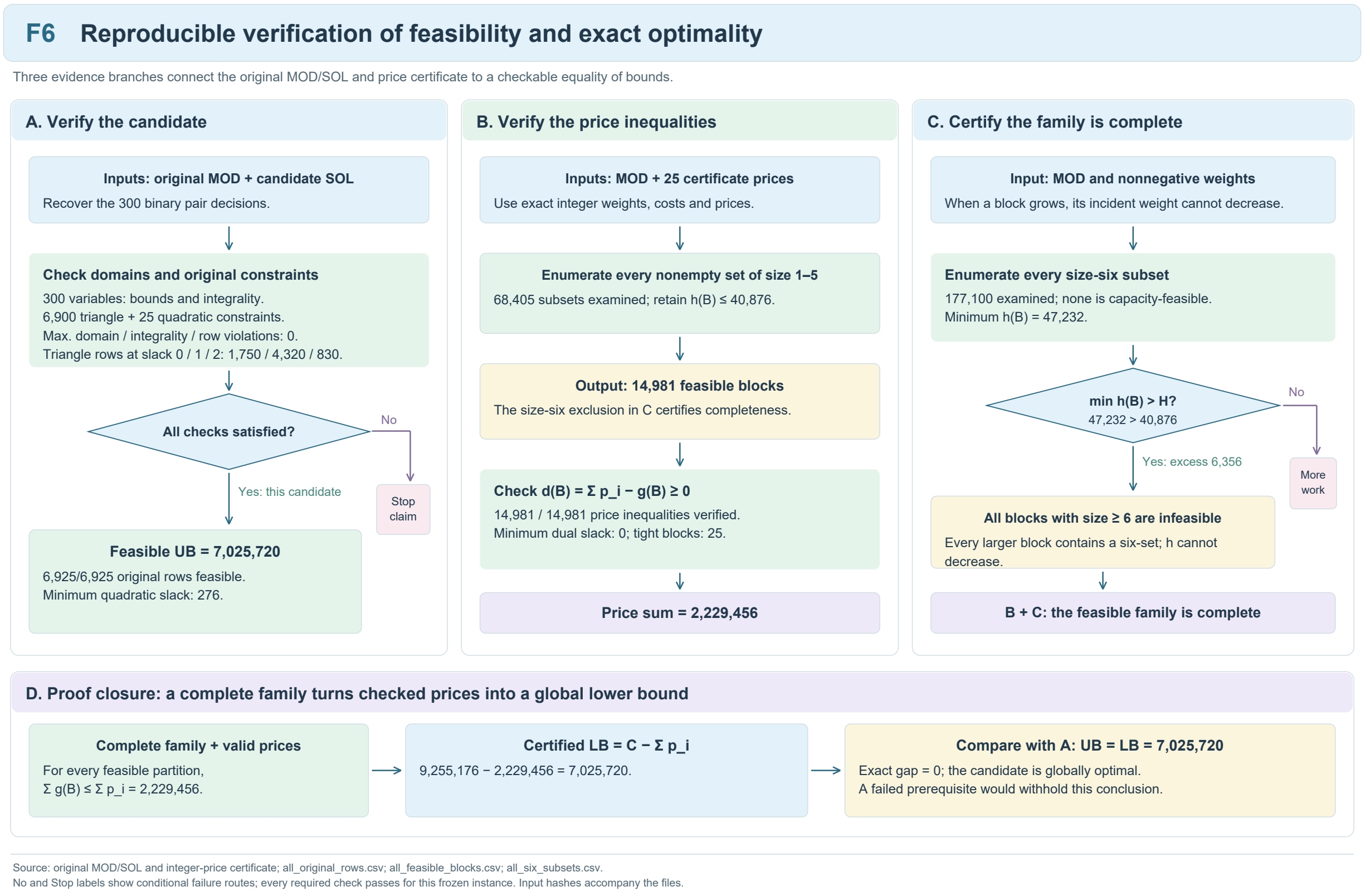

Figure 6. Verification and proof closure. (A) Candidate feasibility. (B) Complete price checks for sizes one to five. (C) Exclusion of every larger block. (D) The complete feasible family and valid prices give a lower bound equal to the feasible upper bound. Conditional failure paths are shown for completeness; all required checks pass for this frozen instance.